\documentclass[conference]{IEEEtran}
\IEEEoverridecommandlockouts
\usepackage{cite}
\usepackage{amsmath,amssymb,amsfonts}
\usepackage{graphicx}
\usepackage{textcomp}
\usepackage{xcolor}
\usepackage{xspace}
\usepackage{xurl}
\def\BibTeX{{\rm B\kern-.05em{\sc i\kern-.025em b}\kern-.08em
    T\kern-.1667em\lower.7ex\hbox{E}\kern-.125emX}}
\usepackage{mathtools}
\usepackage{multirow}
\usepackage{adjustbox}
\usepackage{tikz}
\usetikzlibrary{positioning}
\usepackage{hyperref}
\usepackage{caption}
\usepackage[french,boxed,vlined,linesnumbered,inoutnumbered,rightnl,algo2e]{algorithm2e}
\usepackage{algorithm}
\usepackage{graphicx}
\usepackage{mwe}
\usepackage{stfloats}
\begin{document}

% ---------------------------------------------------------------------------

\title{Integer-only Quantized Transformers for Embedded FPGA-based Time-series Forecasting in AIoT\thanks{The authors gratefully acknowledge the financial support provided by the Federal Ministry for Economic Affairs and Climate Action of Germany for the RIWWER project (01MD22007C).}}
\author{
 \IEEEauthorblockN{Tianheng Ling, Chao Qian, Gregor Schiele}
 \IEEEauthorblockA{Department of Intelligent Embedded Systems, University of Duisburg-Essen, Duisburg, Germany}
 \IEEEauthorblockA{\{tianheng.ling, chao.qian, gregor.schiele@uni-due.de\}}
 }
\maketitle
% ---------------------------------------------------------------------------
\begin{abstract}
This paper presents the design of a hardware accelerator for Transformers, optimized for on-device time-series forecasting in AIoT systems. It integrates integer-only quantization and Quantization-Aware Training with optimized hardware designs to realize 6-bit and 4-bit quantized Transformer models, which achieved precision comparable to 8-bit quantized models from related research. Utilizing a complete implementation on an embedded FPGA (Xilinx Spartan-7 XC7S15), we examine the feasibility of deploying Transformer models on embedded IoT devices. This includes a thorough analysis of achievable precision, resource utilization, timing, power, and energy consumption for on-device inference. Our results indicate that while sufficient performance can be attained, the optimization process is not trivial. For instance, reducing the quantization bitwidth does not consistently result in decreased latency or energy consumption, underscoring the necessity of systematically exploring various optimization combinations. Compared to an 8-bit quantized Transformer model in related studies, our 4-bit quantized Transformer model increases test loss by only 0.63\%, operates up to 132.33$\times$ faster, and consumes 48.19$\times$ less energy. 

\end{abstract}
% -----------------------------

%%%%%%%%%%%%%%%%%%%%%%%%%%%%%%%%%%%%%%%%%%%%%%%%%%%%%%%%%%%%%%%%%%%%%%%%%%%%%%%%%%%%%%%%%%%%%%%%%%%%
\section{Introduction}
\label{sec:introduction}

The integration of Artificial Intelligence with Internet of Things (IoT) devices, commonly referred to as AIoT, is revolutionizing interaction mechanisms with environments, driving innovative solutions in domains such as smart cities and smart homes~\cite{cheng2024advancements}. In these sectors, deploying Deep Learning (DL) models on IoT devices to process sensor data locally offers significant benefits, including reduced data transmission costs and greater independence from network conditions~\cite{liu2023enabling}.

Among DL models, Transformer-based architectures excel in effectively handling long data sequences and capturing global dependencies in fields such as Natural Language Processing (NLP)~\cite{lin2022survey}, Computer Vision (CV)~\cite{khan2022transformers}, and Time-series (TS) analysis~\cite{wen2022transformers}. Despite efforts to simplify these models~\cite{tay2022efficient}, optimized Transformer models are still challenging to deploy on IoT devices due to limited resources and processing power.

To address these challenges, this study aims to refine Transformer models for compact IoT devices while maintaining precision. We adopt a heterogeneous architecture by leveraging embedded Field Programmable Gate Arrays (FPGAs) as hardware accelerators for model inference, specifically targeting time-series forecasting tasks.

Our Transformer model is tailored for the computational constraints of embedded FPGAs by streamlining layers and fine-tuning hyperparameters. Additionally, we employ integer-only quantization and Quantization-Aware Training (QAT) techniques to reduce the numerical bitwidth from 8-bit to 4-bit, optimizing resource utilization and accelerator performance on FPGAs while preserving model precision. Our co-design methodology seamlessly integrates software implementations in PyTorch with optimized hardware components on FPGAs, ensuring smooth accelerator generation and reliable performance. The main contributions of this paper are as follows:

% -----------------------------
\begin{itemize}
    \item We design an FPGA-friendly Transformer model, quantized to 8 bits, that surpasses existing benchmarks by 8.47\% on a traffic flow dataset and by 33.47\% on an air quality dataset. 
    
    \item We provide reusable, resource-optimized, and pipeline-enabled hardware components as VHDL templates, enabling developers to seamlessly translate a trained quantized model with 8, 6, or 4 bits into an FPGA-friendly hardware accelerator without requiring a deep understanding of FPGA development.

    \item We analyze the feasibility of our approach by deploying our generated accelerators on a Xilinx Spartan-7 XC7S15 FPGA. This FPGA is too small to deploy larger variants of our Transformer models, forcing us to compromise between precision and size, similar to a real IoT system. Our results show that we can deploy a 4-bit quantized Transformer model that increases test loss by only 0.63\% but is 132.33$\times$ faster and consumes 48.19$\times$ less energy.
    
\end{itemize}
% -----------------------------

The paper is structured as follows: Section \ref{sec:transformer} explores our FPGA-friendly Transformer architecture. Section \ref{sec:quantization} outlines our approach to integer-only quantization. Section \ref{sec:co_design} details the software-hardware co-design. Section \ref{sec:experiment_results} presents experimental results. Section \ref{sec:related_work} reviews related literature, and Section \ref{sec:conclusion_future_work} concludes the paper and suggests future work.

% ----------------------
\begin{figure}[!htb]
    \centering
    \includegraphics[width=1\columnwidth]{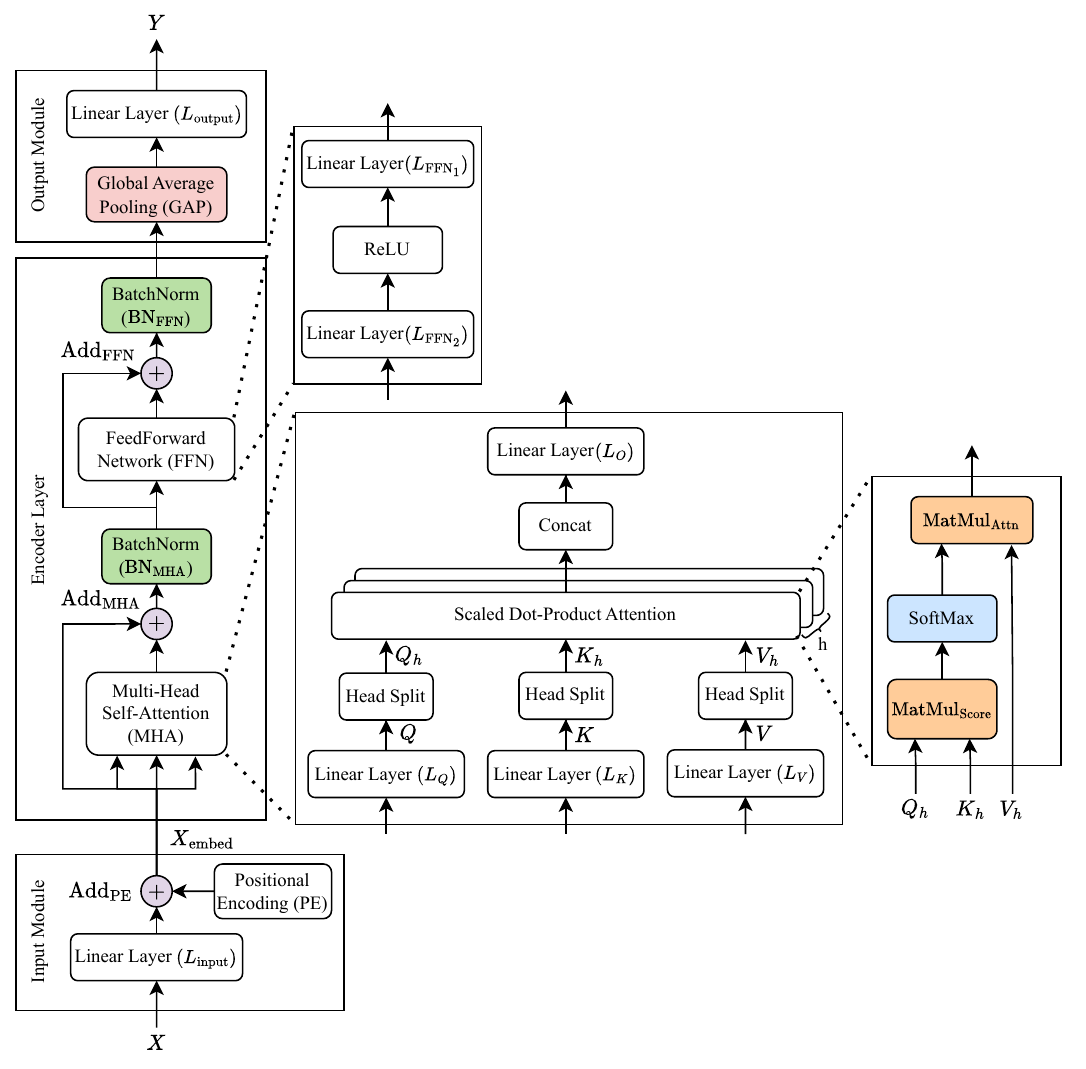}
    \caption{The Architecture of the Transformer Model}
    \label{fig:transformer_model}
\end{figure}
% ----------------------
\begin{table}[!htb]
\footnotesize
\centering
\vspace{-5pt}
\caption{Operations \& Parameters of the Transformer Model}
\label{tab:cal_params}
\begin{adjustbox}{center}
\setlength{\tabcolsep}{1mm}{
\begin{tabular}{|c|c|c|c|c|}
\hline

\multicolumn{2}{|c|}{Modules} & \multicolumn{2}{c|}{Operations} & Parameters \\ \hline

\multicolumn{2}{|c|}{\multirow{3}{*}{Input}} & \multicolumn{2}{c|}{Linear Layer ($L_\text{input}$)} & $(m +1) \times d_\text{model}$ \\ \cline{3-5}
\multicolumn{2}{|c|}{} & \multicolumn{2}{c|}{Positional Encoding (PE)} & - \\ \cline{3-5}
\multicolumn{2}{|c|}{} & \multicolumn{2}{c|}{Addition ($\text{Add}_\text{PE}$)} & - \\ \hline

\multirow{15}{*}{\begin{tabular}[c]{@{}c@{}}Encoder \\ Layer \end{tabular}} & \multirow{7}{*}{MHA} 
& \multicolumn{2}{c|}{Linear Layer ($L_{Q}$)} & $(d_\text{model} + 1) \times d_\text{model}$  \\ \cline{3-5} 
&  & \multicolumn{2}{c|}{Linear Layer ($L_{K}$)}  & $(d_\text{model}+1) \times d_\text{model}$ \\ \cline{3-5} 
&  & \multicolumn{2}{c|}{Linear Layer ($L_{V}$)} & $(d_\text{model}+1) \times d_\text{model}$  \\ \cline{3-5} 
&  & \multirow{3}{*}{\begin{tabular}[c]{@{}c@{}} SDA\end{tabular}}  & MatMul ($\text{MatMul}_\text{Score}$) & - \\\cline{4-5} 
&  &  & Softmax & - \\  \cline{4-5} 
&  &  & MatMul ($\text{MatMul}_\text{Attn}$) & - \\  \cline{3-5} 
&  & \multicolumn{2}{c|}{Output Linear Layer($L_{O}$)}  & $(d_\text{model} + 1) \times d_\text{model}$  \\ \cline{2-5} 
&   \multicolumn{3}{c|}{Addition ($\text{Add}_\text{MHA}$)} & -  \\ \cline{2-5}
&   \multicolumn{3}{c|}{BatchNorm (\(\text{BN}_\text{MHA}\))} & $2 d_\text{model}$ \\  \cline{2-5}

& \multirow{3}{*}{FFN} 
& \multicolumn{2}{c|}{Linear Layer ($L_{\text{FFN}_1}$)} & $4(d_\text{model} + 1) \times d_\text{model}$  \\ \cline{3-5} 
& & \multicolumn{2}{c|}{ReLU}  &  -  \\ \cline{3-5} 
& & \multicolumn{2}{c|}{Linear Layer ($L_{\text{FFN}_2}$)}  &  $(4d_\text{model} + 1) \times d_\text{model}$  \\ \cline{2-5} 
& \multicolumn{3}{c|}{Addition ($\text{Add}_\text{FFN}$)} & - \\ \cline{2-5}  
& \multicolumn{3}{c|}{BatchNorm (\(\text{BN}_\text{FFN}\))} & $2 d_\text{model}$ \\ \hline

\multicolumn{2}{|c|}{\multirow{2}{*}{Output}} & \multicolumn{2}{c|}{Global Average Pooling (GAP)} & -  \\ \cline{3-5} 
\multicolumn{2}{|c|}{} & \multicolumn{2}{c|}{Linear Layer ($L_\text{output}$)}  & $ d_\text{model} + 1$ \\ \hline

\end{tabular}}
\end{adjustbox}
\vspace{-8pt}
\end{table}

% ----------------------

%%%%%%%%%%%%%%%%%%%%%%%%%%%%%%%%%%%%%%%%%%%%%%%%%%%%%%%%%%%%%%%%%%%%%%%%%%%%%%%%
\section{FPGA-friendly Transformer}
\label{sec:transformer}

This section introduces our proposed FPGA-friendly Transformer model for single-step ahead time-series forecasting, adapted from prior work~\cite{ling2023study, becnel2022tiny}. As illustrated in Figure \ref{fig:transformer_model}, the model comprises an input module, an encoder layer, and an output module. It processes the input $X$, a sequence of $n$ data points, each with $m$ dimensions, and is suitable for both univariate and multivariate time-series.

% -----------------------------------------
\subsection{Parameters Simplification}

To simplify the architecture and reduce training complexity, we standardized key dimensions. As detailed in Table \ref{tab:cal_params}, we set the dimensions of the query (Q), key (K), value (V), and output vectors of the Multi-Head Self-Attention (MHA) module to \(d_{\text{model}}\). Based on prior experience, the Feedforward Network (FFN) module dimension is set to four times \(d_{\text{model}}\). Furthermore, we reduced the head count \(h\) in the MHA to 1. Thus, the total number of model parameters is calculated as shown in Equation \ref{eq:total_param}.

% --------------------
\begin{footnotesize}
\vspace{-5pt}
\begin{equation}
\text{Params}_\text{Total} = 12d^2_\text{model} + (15 + m) \times d_\text{model} + 1 
\label{eq:total_param}
\end{equation}
\end{footnotesize}
% --------------------

% -----------------------------------------
\subsection{Encoder Enhancements}
To optimize the encoder layer, we focused on two key enhancements:
% -----------------------------------------
\subsubsection{Scaling Integration}

In conventional Scaled Dot-product Attention (SDA) within the MHA module (See Equation \ref{eq:attention}\footnote{In this paper, the adjacency of two matrices indicates matrix multiplication.}), a scaling factor is applied after matrix multiplication ($\text{MatMul}_\text{Score}$) between Q and the transpose of K (\(K^T\)). This scaling, typically dividing by \(\sqrt{d_\text{model}/h}\), ensures stable gradients during training by keeping the Softmax function within appropriate gradient regions. We directly integrated this scaling operation into the $\text{MatMul}_\text{Score}$ (as detailed in Section \ref{sec:co_design}), thereby eliminating additional computational overhead.

% ----------------------
\begin{footnotesize}
\vspace{-5pt}
\begin{equation}
\text{Attention}(Q, K, V) = \text{Softmax}\left(\frac{Q K^T}{\sqrt{d_\text{model}/h}}\right)V
\label{eq:attention}
\end{equation}
\end{footnotesize}
% ----------------------

% -----------------------------------------
\subsubsection{Batch Normalization}

The original Transformer model proposed in~\cite{ling2023study, becnel2022tiny} utilizes Layer Normalization (LN), which requires the computation of mean and standard deviation during inference. This computation, involving division and square root operations, is computationally expensive for embedded FPGAs. To address this, we replaced LN with Batch Normalization (BN) in our Transformer model, leveraging BN's ability to pre-compute statistics across entire batches during training, thus reducing computational overhead compared to LN's per-feature real-time computation~\cite{liu2023efficient}. Our empirical findings revealed that replacing LN with BN improves precision by 2.78\% and 15.86\% on the selected datasets, respectively, without compromising model performance.

%%%%%%%%%%%%%%%%%%%%%%%%%%%%%%%%%%%%%%%%%%%%%%%%%%%%%%%%%%%%%%%%%%%%%%%%%%%%%%%%%%%%%%%%%%%%%%%%%%%%
\section{Integer-only Quantization}
\label{sec:quantization}

In addition to optimizing the model architecture, we utilize integer-only quantization to reduce data complexity, which is crucial for efficient deployment on embedded FPGAs. As detailed in~\cite{ling2024}, integer-only quantization maps continuous real numbers from the domain \( \mathbb{R} \) to discrete equivalents within a finite set \( \mathbb{Q} \). This process converts a tensor \( X \) to its quantized counterpart \( X_{q} \) using a scale factor \( S \) and a zero point \( Z \). The scale factor \( S \) is a floating-point value, while the zero point \( Z \) is an integer, which is crucial for rounding \( X \) to the nearest integers and clamping the results within the range of \( b \)-bit signed integers \((-2^{b-1}, 2^{b-1}-1)\) as described in Equation~\ref{eq:quantization_mapping}. De-quantization reverts \( X_{q} \) to an approximate real-valued tensor \( X' \), using the same quantization parameters \( S \) and \( Z \), as illustrated in Equation~\ref{eq:dequantization_mapping}. 

% ---------------------------------
\begin{footnotesize}
\vspace{-5pt}
\begin{align}
X \mapsto X_q  & \approx \text{clamp}(\text{round}\left(\frac{X}{S}\right) + Z, -2^{b-1}, 2^{b-1}-1) \label{eq:quantization_mapping} \\
X_q \mapsto X' & = S \cdot (X_q - Z) \label{eq:dequantization_mapping}
\end{align}
\end{footnotesize}
% ---------------

The quantization parameters \( S \) and \( Z \) are dynamically determined during QAT to adapt to the statistical distribution of each tensor~\cite{krishnamoorthi2018quantizing}. This customization enhances model precision with lower-bit quantization compared to Post-training Quantization. Specifically, the scale factor \( S \) and the zero point \( Z \) are computed based on the observed minimum \(\alpha\) and maximum \(\beta\) values of the tensor, as detailed in Equations \ref{eq_asymmetric_signed_scale} and \ref{eq_asymmetric_signed_zero_point}, ensuring accurate representation of the original data distribution.

% -------------------
\begin{footnotesize}
\vspace{-5pt}
\begin{equation}
    S = \frac{\alpha - \beta}{2^b - 1} \label{eq_asymmetric_signed_scale} 
\end{equation}
\end{footnotesize}
% -------------------
\begin{footnotesize}
\vspace{-5pt}
\begin{equation}
    Z \approx \text{clamp}\left(\text{round}((2^{b-1} - 1) - \frac{\alpha}{S}), -2^{b-1}, 2^{b-1}-1\right) \label{eq_asymmetric_signed_zero_point}
\end{equation}
\end{footnotesize}
% --------------------

In this study, all model parameters (as detailed in Table~\ref{tab:cal_params}), along with model inputs, outputs, and inter-layer activations, are designated as quantization objects. This comprehensive quantization ensures that every computation within the model utilizes integer-only operations, facilitating efficient deployment on FPGA platforms. We implemented an asymmetric quantization scheme for all quantization targets, except biases and offsets in BN, which are quantized using a symmetric scheme. While our framework supports mixed-precision quantization, in this paper, we use a uniform quantization bitwidth across all quantization objects for simplicity and consistency.

%%%%%%%%%%%%%%%%%%%%%%%%%%%%%%%%%%%%%%%%%%%%%%%%%%%%%%%%%%%%%%%%%%%%%%%%%%%%%%%%
\vspace{-2 pt}
\section{Software-Hardware Co-Design}
\label{sec:co_design}

A software-hardware co-design is essential for deploying the Transformer model on resource-constrained FPGAs. Our approach merges integer-only quantization with custom hardware optimizations at the register transfer level (RTL), creating VHDL templates that enhance computational efficiency and reduce resource utilization. We adapted the linear layer and ReLU components from prior work~\cite{ling2024} on Multilayer Perceptron to handle multiple input dimensions. By processing these dimensions sequentially rather than duplicating logic circuits, we avoid excessive resource consumption on FPGAs. The following subsections detail the co-design for Transformer-specific operations.

% ----------------------------------------------------------
\vspace{-5pt}
\subsection{Addition Component}

The addition operation is described in Equation \ref{eq:add_step0}, where two floating-point inputs, \(A^{1}\) and \(A^{2}\), are added to produce a floating-point output \(A^{3}\). In our Transformer model, there are three addition instances labeled $\text{Add}_\text{PE}$ to $\text{Add}_\text{FFN}$ in Table \ref{tab:cal_params}. For integer-only addition, as outlined in Section~\ref{sec:quantization}, \(A^{3}\) can be approximated using Equation \ref{eq:add_step1}. To obtain the integer output \(A^{3}_q\), Equation \ref{eq:add_step1} is transformed into Equation \ref{eq:add_step2}. The terms \( \frac{S_{A^{1}}}{S_{A^{3}}} \) and \( \frac{S_{A^{2}}}{S_{A^{3}}} \) are the remaining floating-point values, which can be approximated using a precomputed positive integer \(M\) by right-shifting by \(n\) bits, as shown in Equation \ref{eq:add_step4} (using \( \frac{S_{A^{1}}}{S_{A^{3}}} \) as an example). This process, referred to as \textit{ApproxMul}, ensures that all operations remain within the integer domain. Notably, the $\text{Add}_\text{PE}$ operation in the input module uses precomputed positional information from a look-up table.

% -------------------
\begin{footnotesize}
\vspace{-3pt}
\begin{equation}
A^{3} = A^{1} + A^{2} \label{eq:add_step0} \\
\end{equation}
\end{footnotesize}
% -------------------
\begin{footnotesize}
\vspace{-8pt}
\begin{equation}
    A^{3} \approx S_{A^{1}} \cdot (A^{1}_{q}-Z_{A^{1}}) + S_{A^{2}} \cdot (A^{2}_{q}-Z_{A^{2}}) \label{eq:add_step1} \\
\end{equation}
\end{footnotesize}
% -------------------
\begin{footnotesize}
\vspace{-8pt}
\begin{equation}
    A^{3}_{q} \approx \left(\frac{S_{A^{1}}}{S_{A^{3}}} \cdot (A^{1}_{q}-Z_{A^{1}}) + \frac{S_{A^{2}}}{S_{A^{3}}} \cdot (A^{2}_{q}-Z_{A^{2}})\right) + Z_{A^{3}} \label{eq:add_step2} \\
\end{equation}
\end{footnotesize}
% -------------------
\begin{footnotesize}
\vspace{-8pt}
\begin{equation}
\frac{S_{A^{1}}}{S_{A^{3}}} \approx 2^{-n} \cdot M \label{eq:add_step4}
\end{equation}
\end{footnotesize}
% -------------------

% ----------------------------------------------------------
\subsection{Matrix Multiplication Component}

The co-design of the matrix multiplication component follows similar principles to the linear layer component but replaces static weights with dynamic inputs and excludes bias terms. This component involves two instances, denoted as $\text{MatMul}_\text{Score}$ and $\text{MatMul}_\text{Attn}$, in Table \ref{tab:cal_params}. As mentioned in Section \ref{sec:transformer}, $\text{MatMul}_\text{Score}$ computes the dot product between matrices \(Q\) and \(K^T\). We integrate the scaling factor \( \sqrt{d_{K}/h} \) into the quantization scaling factor (see Equation~\ref{eq:matmul}), eliminating the need for adjustments at the RTL level and streamlining the hardware implementation process.

% -------------------
\begin{footnotesize}
\begin{equation}
A^3_{q} \approx \frac{S_{A^1} \cdot S_{A^2}}{S_{A^3} \cdot \sqrt{d_\text{model}/h} } \left((A^1_{q}-Z_{A^1})(A^2_{q}-Z_{A^2}) \right) + Z_{A^3}
\label{eq:matmul}
\end{equation}
\end{footnotesize}
% -------------------

While matrix transposition of $K$ is easily handled in the PyTorch framework with a call to \verb|transpose()|\footnote{\UrlFont{https://pytorch.org/docs/stable/generated/torch.transpose.html}}, it poses significant challenges on resource-constrained FPGAs due to its high time and memory requirements. To mitigate these costs, we implement an address-mapping mechanism that enables direct data retrieval from the non-transposed input buffer, avoiding extra time and memory for matrix transposition. This component also serves $\text{MatMul}_{Attn}$ by deactivating the address mapping block.
% ---------------------------------------------------------------------------

\subsection{Softmax Component}

Although many quantization approaches for the Softmax function have been proposed~\cite{stevens2021softermax, kim2021bert, li2023vit}, our experiments revealed that none could meet our precision criteria for regression tasks. Inspired by prior work~\cite{vasyltsov2021efficient}, we adopt a Softmax implementation based on lookup tables for computing \verb|exp()| for all possible inputs. This approach relies on two lookup tables: \textit{NLUT} for numerators and \textit{DLUT} for denominators. As is customary, we constrain the output of the \verb|exp()| function to the interval (0, 1] by offsetting all potential integer inputs \(X_{q}\) by the maximum value, yielding \(\hat{X}_{q}\). Subsequently, the exponential values \(E\) are derived through \(E=\exp(\hat{X}_{q})\). To ensure effective quantization, we introduce a scaling factor \(S_E\) and a zero point \(Z_E\) as defined in Equations~\ref{eq:quant_softmax1} and~\ref{eq:quant_softmax11}, accommodating the dynamic range within the constraints of integer precision. The \textit{DLUT} and \textit{NLUT} are calculated using Equations~\ref{eq:quant_softmax2} and \ref{eq:quant_softmax3}, respectively, ensuring bitwidths of $2b$ for \textit{DLUT} and $3b$ for \textit{NLUT} to maintain sufficient precision. The integer output \(A_q\) is computed by retrieving the quantized numerators and denominators and performing the division as outlined in Equation~\ref{eq:quant_softmax4}, where \(i,j \in [1, n]\).

% -------------------------
\begin{footnotesize}
    \begin{equation}
        S_E = \frac{1}{((2^{(2b - 1)}-1) - (-2^{(2b - 1)})) / (n^2 \cdot h)} \label{eq:quant_softmax1} \\
    \end{equation}
\end{footnotesize}
% -------------------------
\begin{footnotesize}
    % \vspace{-5pt}
    \begin{equation}
        Z_E = 2^{(2b - 1)} - \frac{1}{S_E} \label{eq:quant_softmax11} \\
    \end{equation}
\end{footnotesize}
% -------------------------
\begin{footnotesize}
    \vspace{-5pt}
    \begin{equation}
        \text{DLUT}(\hat{X}_q) = \text{clamp}(\text{round}(\frac{E}{S_E}), -2^{2b-1}, 2^{2b-1}-1 ) \label{eq:quant_softmax2} \\
    \end{equation}
\end{footnotesize}
% -------------------------
\begin{footnotesize}
    % \vspace{-3pt}
    \begin{equation}
        \text{NLUT}(\hat{X}_q) = \text{clamp}(\text{round}(\frac{E}{S_E \cdot S_A}),-2^{3b-1}, 2^{3b-1}-1) \label{eq:quant_softmax3} \\
    \end{equation}
\end{footnotesize}
% -------------------------
\begin{footnotesize}
    % \vspace{-3pt}
    \begin{equation}
        A_q \approx \frac{\text{NLUT}(\hat{X}_q(i,j))}{\sum^{n}_{i=1} (\text{DLUT}(\hat{X}_q(i,j)) - Z_E)} + Z_A \label{eq:quant_softmax4}
    \end{equation}
\end{footnotesize}
% -------------------------

Our Softmax implementation on FPGA involves three phases, as detailed in Algorithm~\ref{alg:softmax}. In the first phase (Lines 2-5), the algorithm iteratively scans each row to identify the maximum value (\(\text{max}_i\)). In the second phase (Lines 6-11), all inputs are normalized by subtracting \(\text{max}_i\). The resulting \(x_q\) values are then passed through \textit{NLUT} and \textit{DLUT} to prepare the numerators and denominators, respectively. The outputs from \textit{NLUT} are stored in \(X_\text{numerators}\), while those from \textit{DLUT} are summed up. In the third phase (Lines 12-13), element-wise division is performed to produce the final output.

However, this division operation is resource-intensive on FPGAs. Our evaluation of the default divider in Vivado synthesis revealed a 23 ns logic latency, limiting the system's clock frequency to below 43 MHz. Recognizing that higher clock frequencies enhance the energy efficiency of Spartan-7 FPGAs~\cite{qian2023energy}, we employ a \textit{Radix-2} non-restoring divider~\cite{chiang1999radix}, enabling the system to operate at higher clock frequencies. Although each division operation may require more clock cycles, it constitutes only a small portion of the total cycles required for one model inference. Thus, this optimization can significantly reduce overall inference time.

% -------------------------

\begin{minipage}{.95\columnwidth}
\centering
\begin{algorithm}[H]
\footnotesize
\caption{Softmax Implementation on FPGA}
\label{alg:softmax}
\KwIn{$X_q$ of size $n \times n$} 

\For{$i \gets 1$ \KwTo $n$}{
    
    $\text{max}_i \gets -\infty$ 
    
    \For{$j \gets 1$ \KwTo $n$} {
        \If{$X_q[i][j] > \text{max}_i$}{
            $\text{max}_i \gets X_q[i][j]$
     }
}
     
    $\text{sum} \gets 0$ 
    
    \For{$j \gets 1$ \KwTo $n$} {
    
    $ x_q \gets X_q[i][j] - \text{max}_i$ 
    
    $ X_{\text{numerators}}[j] \gets \text{NLUT}(x_q) $ 
    
    $ x_{\text{denominator}} \gets \text{DLUT}(x_q) $ 
    
    $ \text{sum} \gets sum + (x_\text{denominator} - Z_E) $
    }

    \For{$j \gets 1$ \KwTo $n$} {
    
    $A_q[i][j] \gets \frac{X_{\text{numerators}}[j]}{\text{sum}} + Z_A$
    }
    
}
\KwOut{$A_q$ of size $n \times n$}

\end{algorithm}
\end{minipage}

% ---------------------------------------------------------------------------

\subsection{Batch Normalization Component}

BN operates on input \(X\) with dimensions \((n, d_\text{model})\) to normalize the data using Equation~\ref{eq:quant_bn1}, where \( \mu_j \) and \( \sigma_j^2 \) represent the mean and variance of feature \( j \in [1, d_\text{model}] \) across all samples in the batch. The scaling factor \( \gamma_j \) and offset \( \beta_j \) adjust the normalized values, while \( \epsilon \) prevents division by zero. By transforming, Equation~\ref{eq:quant_bn1} can be expressed as Equation \ref{eq:quant_bn2} with new a scaling factor \(\hat{\gamma_{j}}\) (equals to \(\frac{\gamma_{j}}{\sqrt{\sigma^2_{j} + \epsilon}}\)), and a new offset \(\hat{\beta}_{j}\) (represents \(\beta_j - \frac{\mu_j}{\sqrt{\sigma_j^2 + \epsilon}}\)). Following a similar principle adopted in linear layer component, the integer output \(A_q(i,j)\) can be obtained using Equation~\ref{eq:quant_bn3}. Notably, we approximate the offset term \( S_{\hat{\beta}_{j}}\cdot\hat{\beta}_{j_q} \)as \(S_{\hat{\gamma_{j}}} \cdot S_X \cdot \hat{\beta^*}_{j_q} \) to streamline the calculation. Thus, the hardware implementation focuses on designing an efficient pipeline to perform element-wise dot products concurrently during data fetching. The newly computed products are then scaled using the \textit{ApproxMul} operation to obtain the integer output.

% -----------------
\begin{footnotesize}
\vspace{-5pt}
\begin{align}
A(i,j) = \gamma_j \cdot \frac{X(i,j) - \mu_j}{\sqrt{\sigma_j^2 + \epsilon}} + \beta_j 
\label{eq:quant_bn1} 
\end{align}
\end{footnotesize}
% -----------------
\begin{footnotesize}
\vspace{-10pt}
\begin{align}
A(i,j) = \hat{\gamma_{j}} \cdot X(i,j) + \hat{\beta}_{j}
\label{eq:quant_bn2}
\end{align}
\end{footnotesize}
% -----------------
% \begin{footnotesize}
% \vspace{-10pt}
% \begin{align}
% S_A \cdot (A_q(i,j)-Z_A)  & \approx \hat{\gamma_{j}} \cdot S_X \cdot (X_q(i,j)-Z_X) + \hat{\beta}_{j}\label{eq:quant_bn3} 
% \end{align}
% \end{footnotesize}
% -----------------
\begin{footnotesize}
\vspace{-10pt}
% \begin{align}
% A_q(i,j)   & \approx \frac{S_{\hat{\gamma_{j}}} \cdot S_X}{S_A} \cdot ((\hat{\gamma_{j}}_q - Z_{\hat{\gamma_j}})(X_q(i,j)-Z_X) + \hat{\beta^*_{j_q}}) + Z_A \label{eq:quant_bn3}
% \end{align}
\begin{align}
A_q(i,j) &\approx 
\frac{S_{\hat{\gamma_j}} \, S_X}{S_A}
\Bigl[ \bigl(\hat{\gamma}_{j_q} - Z_{\hat{\gamma_j}}\bigr)
       \bigl(X_q(i,j) - Z_X\bigr) 
       + \hat{\beta}_{j_q}^* \Bigr] 
+ Z_A
\label{eq:quant_bn3}
\end{align}
\end{footnotesize}
% -----------------
% \begin{footnotesize}
% \vspace{-5pt}
% \begin{align}
% S_A \cdot (A_q(i,j)-Z_A)  & \approx \hat{\gamma_{j}} \cdot S_X \cdot (X_q(i,j)-Z_X) + \hat{\beta}_{j}\label{eq:quant_bn3} 
% \end{align}
% \begin{align}
% A_q(i,j)   & \approx \frac{S_{\hat{\gamma_{j}}} \cdot S_X}{S_A} \cdot (\hat{\gamma_{j}}_q - Z_{\hat{\gamma_j}})((X_q(i,j)-Z_X) + \hat{\beta^*_{j_q}}) + Z_A \label{eq:quant_bn4}
% \end{align}
% \end{footnotesize}
% -----------------

% ---------------------------------------------------------------- 
\subsection{Global Average Pooling Component}

Given the input \(X\) with dimensions \((n, d_\text{model})\), the GAP operation computes the average over dimension \(n\), producing the quantized output \(A_q\), as shown in Equation~\ref{eq:gap}. To circumvent division operations and considering the fixed sequence length \(n\), we incorporate the division factor \(1/n\) into the quantization scaling factor. Thus, the FPGA implementation requires only a summation operation followed by an \textit{ApproxMul} operation.

% ---------------------
\begin{footnotesize}
\vspace{-5pt}   
\begin{equation}
A_{q}(j) = \frac{S_{X}}{S_{A} \cdot n} \cdot \sum_{i=1}^{n}(X_{q}(i, j)-Z_{X}) + Z_{A}, \quad j \in [1, d_\text{model}] \label{eq:gap}
\end{equation}  
\end{footnotesize}
% ---------------------

%%%%%%%%%%%%%%%%%%%%%%%%%%%%%%%%%%%%%%%%%%%%%%%%%%%%%%%%%%%%%%%%%%%%%%%%%%%%%%%%
\section{Experiments and Results}
\label{sec:experiment_results}

This section outlines our experimental setup, including dataset details and processing methods. We discuss model precision, resource utilization, inference time, and power and energy consumption on a Spartan-7 XC7S15 FPGA across various model configurations.

% -------------------------------------------------
\subsection{Datasets and Data Processing}
  
We used two datasets for our case studies: the \textit{PeMS}\footnote{\UrlFont{https://doi.org/10.5281/zenodo.3939793}} dataset, which captures univariate traffic flow data from 11,160 sensor measurements over four weeks. Each series samples data at 5-minute intervals, yielding 8,064 time points. We selected a single series with sensor index 4192 to facilitate a fair comparison with~\cite{qian2023energy}. The \textit{AirU}\footnote{\UrlFont{https://dx.doi.org/10.21227/aeh2-a413}} dataset contains multivariate air quality records from 19,380 observations, featuring seven variables with Ozone as the target variable. After removing discontinuous records, the dataset was reduced to 15,258 feature-target pairs. Observations overlapping with the test set period used in \cite{becnel2022tiny} were split into 14,427 training samples and 831 testing samples. All data were normalized using the MinMax method to ensure uniformity in training and testing inputs.

% -------------------------------------------------
\subsection{Experiments Setup}

Each model configuration underwent 50 training sessions consisting of 100 epochs, with early stopping implemented to prevent overfitting. We used a batch size of 256 and the Adam optimizer with standard parameters ($\beta_1 \!=\! 0.9$, $\beta_2 \!=\! 0.98$, $\epsilon \!=\! 10^{-9}$). The learning rate was initialized at 0.001, with a scheduler having a step size of 3 and a decay factor $\gamma$ of 0.5 for dynamic adjustment during training. These training sessions were conducted on an NVIDIA GeForce RTX 2080 SUPER GPU, utilizing CUDA 11.0 and PyTorch 3.11 within the Ubuntu 20.04.6 LTS. The objective metric for training was the minimization of the Mean Squared Error. The Mean Squared Loss function guided the training process. Post-training, we applied an inverse transformation to the model’s outputs and normalized target values to compute the Root Mean Square Error (RMSE) on the test data as the evaluation metric. 
To generate the accelerator, Python scripts were used to translate the trained quantized model by passing its model and quantization parameters to VHDL templates, which resulted in corresponding VHDL files. We then used GHDL simulations to estimate the number of clock cycles required per inference. The accelerator design was synthesized using Vivado, generating comprehensive reports on resource utilization, timing, and power consumption. Finally, the accelerators were validated on real hardware to assess the effectiveness and efficiency of our FPGA implementation. Relevant source code is provided in the accompanying GitHub repository\footnote{\url{https://github.com/tianheng-ling/TinyTransformer4TS}}.

% -----------------------------------------------------
\begin{table*}[tp] 
\footnotesize
\centering
\caption{Parameter Count and Test RMSE Across Various Models Configurations}
\label{tab:model_precision}
\begin{adjustbox}{center}
\setlength{\tabcolsep}{1.3mm}{
\begin{tabular}{|c|c|c|c|c|c|c|c|c|c|c|c|}
\hline
\multicolumn{2}{|c|}{\multirow{2}{*}{Configs.}} & \multicolumn{5}{c|}{PeMS} & \multicolumn{5}{c|}{AirU}   \\ \cline{3-12} 
\multicolumn{2}{|c|}{}  & \multicolumn{1}{c|}{\multirow{2}{*}{Params}} & \multicolumn{4}{c|}{RMSE} & \multicolumn{1}{c|}{\multirow{2}{*}{Params}} & \multicolumn{4}{c|}{RMSE} \\ \cline{1-2} \cline{4-7} \cline{9-12} 
\multicolumn{1}{|c|}{n} & $d_\text{model}$ & \multicolumn{1}{c|}{} & \multicolumn{1}{c|}{FP32} & \multicolumn{1}{c|}{8-bit} & \multicolumn{1}{c|}{6-bit} & 4-bit & \multicolumn{1}{c|}{} & \multicolumn{1}{c|}{FP32} & \multicolumn{1}{c|}{8-bit} & \multicolumn{1}{c|}{6-bit} & 4-bit \\ \hline 
\multirow{4}{*}{6}  
&   8 & 897  & 0.1609 & 0.1809 & 0.2227 & 0.4224 & 945  & 4.055 & 4.791 & 4.696 & 6.616 \\ \cline{2-12}
&  16 & 3329  & 0.1615 & 0.1685 & 0.1988 & \textbf{0.3117} & 3425  & 3.726 & 3.858 & 4.004 & 6.566 \\ \cline{2-12}
&  32 & 12801 & 0.1652 & \textbf{0.1563} & 0.1971 & 0.3300 & 12993 & 3.799 & 3.665 & 4.095 & 5.731 \\ \cline{2-12}
&  64 & 50177 & 0.1666 & 0.1621 & 0.1855 & 0.4084 & 50561 & 4.236 & \textbf{3.506} & 3.923 & 5.699 \\ \hline

\multirow{4}{*}{12} 
&   8 & 897  & 0.1692 & 0.1847 & 0.2107 & 0.4226 & 945  & 3.846 & 5.274 & 6.229 & 6.406 \\ \cline{2-12}
&  16 & 3329  & 0.1582 & 0.1596 & 0.1922 & 0.3979 & 3425  & 3.878 & 4.804 & 4.675 & 5.853 \\ \cline{2-12}
&  32 & 12801 & 0.1568 & 0.1617 & 0.1848 & 0.4030 & 12993 & 3.859 & 3.723 & 4.116 & 5.474 \\ \cline{2-12}
&  64 & 50177 & 0.1586 & 0.1571 & \textbf{0.1809} & 0.3279 & 50561 & 4.071 & 3.518 & \textbf{3.763} & 5.494 \\ \hline

\multirow{4}{*}{18} 
&   8 & 897  & 0.1645 & 0.2008 & 0.2158 & 0.5258  & 945  & 3.917 & 5.419 & 6.757 & 6.765 \\ \cline{2-12}
&  16 & 3329  & 0.1593 & 0.1607 & 0.1838 & 0.3297  & 3425  & 3.662 & 4.974 & 4.614 & 5.866 \\ \cline{2-12}
&  32 & 12801 & \textbf{0.1567} & 0.1581 & 0.1918 & 0.3162 & 12993 & \textbf{3.603} & 3.796 & 4.012 & 5.795 \\ \cline{2-12}
&  64 & 50177 & 0.1600 & 0.1583 & 0.1845 & 0.3462 & 50561 & 4.055 & 3.518 & 3.897 & \textbf{5.286} \\ \hline

\multirow{4}{*}{24} 
&   8  & 897  & 0.1664 & 0.2229 & 0.2436 & 0.4841 & 945 & 4.129 & 5.377 & 7.160 & 6.795 \\ \cline{2-12}
&  16  & 3329  & 0.1595 & 0.1673 & 0.1988 & 0.3856 & 3425 & 3.899 & 4.929 & 5.776 & 5.881 \\ \cline{2-12}
&  32  & 12801 & 0.1574 & 0.1591 & 0.1943 & 0.4158 & 12993 & 3.867 & 4.019 & 4.380 & 5.700\\ \cline{2-12}
&  64  & 50177 & 0.1604 & 0.1573 & 0.1874 & 0.3787 & 50561 & 3.988 & 3.619 & 3.840 & 5.456 \\ \hline
\end{tabular}
}
\end{adjustbox}
\vspace{-10pt}
\end{table*}
% -----------------------------------------------------

% -------------------------------------------------
\subsection{Model Precision Across Different Configurations}
\label{subsec:model_precision}

To investigate the impact of model complexity on precision, we conducted experiments on both datasets with varying numbers of input features (\(m\)). These experiments also explored different input lengths (\(n\): 6, 12, 18, 24) and embedding dimensions (\(d_{\text{model}}\): 8, 16, 32, 64), evaluating the model at different precision levels: floating-point numbers (FP32), 8-bit integers (8-bit), 6-bit integers (6-bit), and 4-bit integers (4-bit).

% -------------------------------------------------
\subsubsection{Parameter Count and FP32 Models}

The \emph{Params} column in Table~\ref{tab:model_precision} displays the parameter count of each model. The variance in the number of input features (\(m\)) between datasets did not significantly affect the overall parameter count, as \(m\) only influences the parameter count of the linear layer \(L_{input}\). Increasing the embedding dimension (\(d_{\text{model}}\)) led to a notable rise in model parameters. However, this increase did not consistently improve the precision of FP32 models, indicating that larger \(d_{\text{model}}\) does not necessarily enhance performance on these datasets. Additionally, increasing the input length (\(n\)) did not consistently improve precision, suggesting that incorporating more historical data points does not linearly enhance precision.

Overall, the optimal configuration for minimizing test RMSE across both datasets was \(n\!=\!18\) and \(d_{\text{model}}\!=\!32\). Notably, our FP32 model outperformed benchmarks reported in~\cite{qian2022enhancing} for the PeMS dataset, demonstrating a 16.06\% improvement. Similarly, compared to FP32 results on the AirU dataset documented by Becnel et al.~\cite{becnel2022tiny} (where \(n\!=\!24\) and \(d_{\text{model}}\!=\!64\)), our model achieved a 3.20\% enhancement.

% -------------------------------------------------
\subsubsection{Quantization and Model Precision}

\begin{figure}[!htb]
    \vspace{-5pt}
    \centering
    \includegraphics[width=1\columnwidth]{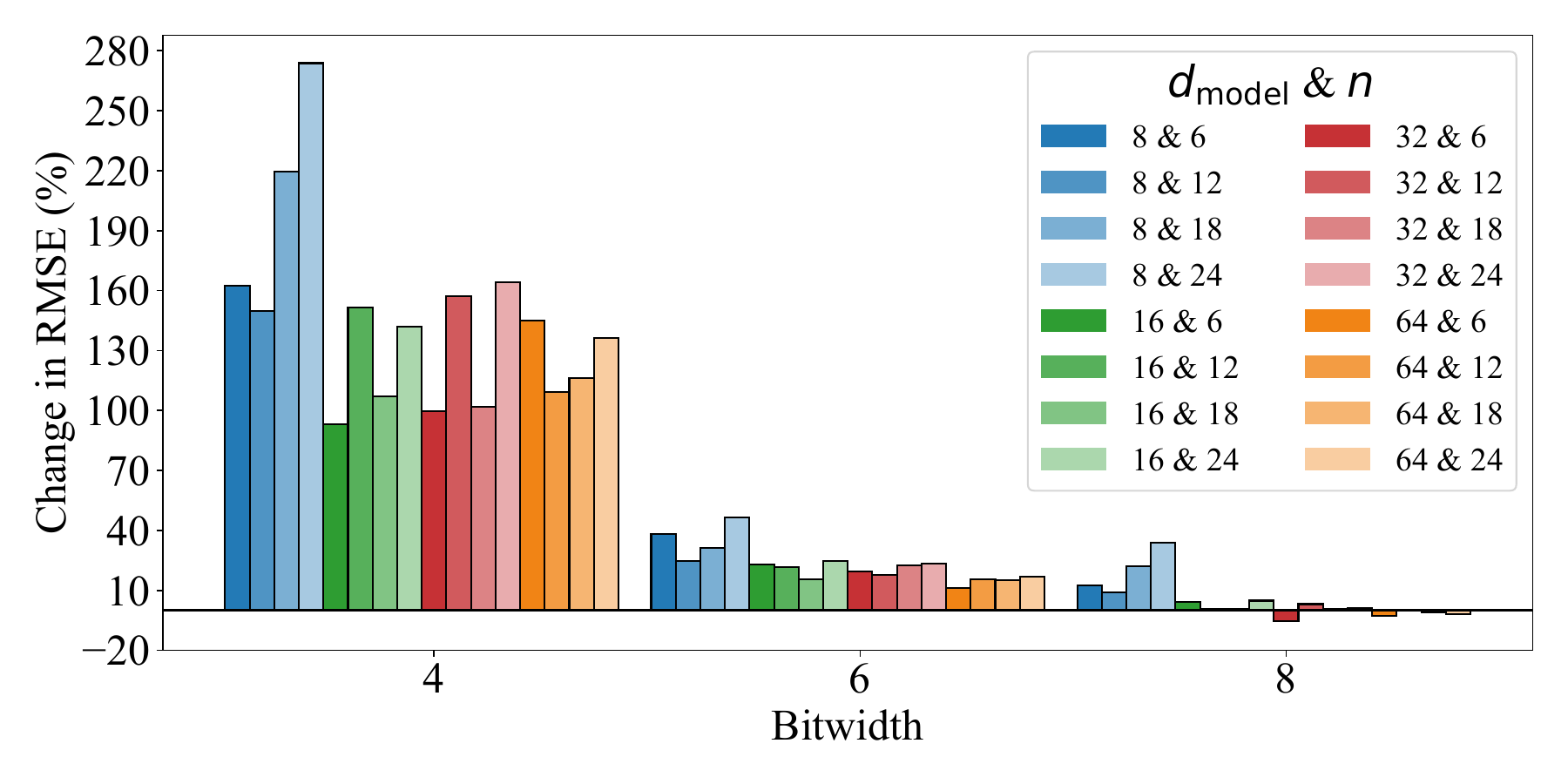}
    \vspace{-10pt}   
    \caption{PeMS Dataset: RMSE Variation}
    \label{fig:quant_pems_comparison}
    \vspace{-10pt}
\end{figure}

\begin{figure}[!htb]
    \vspace{-10pt}
    \centering
    \includegraphics[width=1\columnwidth]{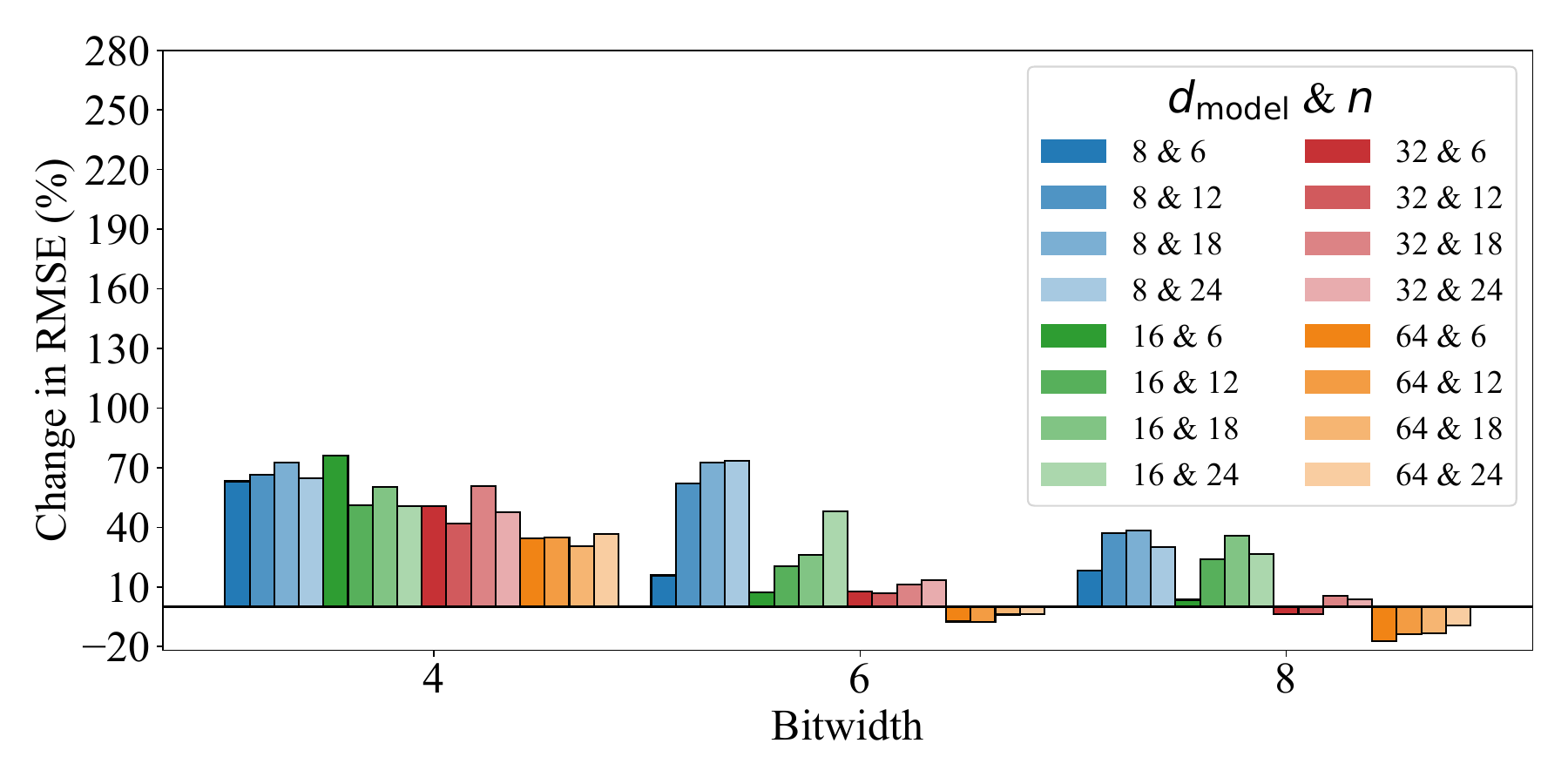}
    \vspace{-10pt}   
    \caption{AirU Dataset: RMSE Variation}
    \label{fig:quant_airU_comparison}
    \vspace{-5pt}   
\end{figure}

Figures~\ref{fig:quant_pems_comparison} and \ref{fig:quant_airU_comparison} depict the RMSE variation of quantized models compared to FP32 models on the PeMS and AirU datasets, respectively. Smaller \(d_{\text{model}}\) models exhibited greater sensitivity to quantization across various datasets and bitwidths, as indicated by the taller blue bars in the figures. This suggests that higher-dimensional embeddings better preserve essential information even with reduced numerical precision. However, the impact of changes in \(n\) (input length) on model precision and quantization sensitivity remained uncertain.

% ------------------------------
% ------------------------------
% ------------------------------

The increase (at least 93.0\%) in RMSE at 4-bit quantization is more pronounced on the PeMS dataset than on the AirU dataset, likely due to its univariate nature. In contrast, the multivariate nature of the AirU dataset provides enhanced robustness against quantization effects, demonstrating up to a 76\% increase in RMSE at 4-bit quantization. Notably, at \(d_{\text{model}} \!=\! 64\), the 8-bit and 6-bit quantized models on the AirU dataset exhibited up to 17.23\% and 7\% lower RMSE, respectively, compared to their FP32 counterparts. On the PeMS dataset, however, only the 8-bit quantized model achieved 5.38\% lower RMSE compared to FP32.

The optimal configuration of the FP32 model did not consistently yield the best precision under quantization. For instance, on the AirU dataset, the optimal configuration (with \(n\!=\!6\) and \(d_{\text{model}}\!=\!64\)) under 8-bit quantization resulted in 2.69\% lower RMSE compared to the best FP32 model. Under 6-bit quantization, it (with \(n\!=\!12\) and \(d_{\text{model}}\!=\!64\)) had 4.44\% higher RMSE than the best FP32 model. For 4-bit quantization, \(n\!=\!18\) and \(d_{\text{model}}\!=\!64\) provided the best outcomes, with RMSE 46.71\% higher than the best FP32 model. Notably, on the PeMS dataset, our 8-bit quantized model outperformed the 8-bit quantized model reported in~\cite{qian2022enhancing} by 8.47\%. Even our 6-bit quantized model delivered comparable performance. In addition, on the AirU dataset, our 8-bit quantized model surpassed the 8-bit quantized model reported in~\cite{becnel2022tiny} by 33.47\%. Remarkably, even our 4-bit quantized model outperformed their 8-bit counterpart by 2.83\%.

% ---------------------------------------------------------------------------
\vspace{-5pt}
\subsection{Resource Utilization}
\label{subsec:resource_util}

Due to space constraints, this section focuses on the AirU dataset and evaluates resource utilization, including Lookup Tables (LUTs), Block RAM (BRAM), and Digital Signal Processing Slices (DSPs), for various model configurations and quantization bitwidths on the XC7S15 FPGA. As illustrated in Table~\ref{tab:model_resource}, even when quantized to 4-bit, the largest configuration (\(n\!=\!24\) and \(d_{\text{model}}\!=\!64\)) exceeded the FPGA's resource capacity, hence could not be deployed (denoted by ``-''). Conversely, the smallest configuration (\(n\!=\!6\) and \(d_{\text{model}}\!=\!8\)) fit easily within the FPGA's constraints across different quantization bitwidths, with consistent BRAM utilization at 10\%. LUTs and DSPs utilization increase with bitwidth, indicating higher logic resource demand with greater numeric precision.

% -----------------------------------------------------

\begin{table}[!htb] \footnotesize
\centering
\caption{Resource Utilization on AirU Dataset}
\label{tab:model_resource}
\begin{adjustbox}{center}
\setlength{\tabcolsep}{1.2mm}{
\begin{tabular}{|c|c|c|c|c|c|c|c|c|c|c|}
\hline

\multicolumn{2}{|c|}{Configs.} & \multicolumn{9}{c|}{AirU}  \\ \hline

\multicolumn{1}{|c|}{\multirow{2}{*}{n}} & \multirow{2}{*}{$d_\text{model}$} & \multicolumn{3}{c|}{LUTs(\%)} & \multicolumn{3}{c|}{BRAM(\%)}  & \multicolumn{3}{c|}{DSPs(\%)} \\ \cline{3-11} 

\multicolumn{1}{|c|}{}   &   & \multicolumn{1}{c|}{8-bit} & \multicolumn{1}{c|}{6-bit} & \multicolumn{1}{c|}{4-bit} & \multicolumn{1}{c|}{8-bit} & \multicolumn{1}{c|}{6-bit} & \multicolumn{1}{c|}{4-bit} & \multicolumn{1}{c|}{8-bit} & \multicolumn{1}{c|}{6-bit} & 4-bit \\ \hline

\multirow{4}{*}{6}  
&   8 & 55.6 & 42.0 & 34.2 & 10.0 & 10.0 & 10.0 & 100.0 & 90.0 & 65.0   \\ \cline{2-11}
&  16 & 57.1 & 47.0 & 37.3 & 40.0 & 30.0 & 30.0 & 100.0 & 95.0 & 65.0 \\ \cline{2-11}
&  32 & 62.7 & 50.5 & 41.1 & 55.0  & 55.0 & 40.0 & 100.0 & 95.0 & 65.0  \\ \cline{2-11}
&  64 & \textbf{89.5} & 57.2 & 47.4 & \textbf{100.0} & 100.0 & 75.0 & \textbf{100.0}& 90.0 & 60.0 \\ \hline

\multirow{4}{*}{12} 
&   8 & 58.0 & 46.0 & 36.5 & 20.0 & 10.0 & 10.0 & 100.0 & 95.0 & 65.0  \\ \cline{2-11}
&  16 & 65.1 & 51.5 & 42.7 & 45.0  & 35.0 & 35.0 & 100.0 & 95.0 & 60.0 \\ \cline{2-11}
&  32 & 74.3 & 58.9 & \textbf{46.8} & 60.0 & 55.0 & \textbf{40.0} & 100.0 & 95.0 & \textbf{65.0} \\ \cline{2-11}
&  64 & - & \textbf{75.3} & 58.2 &  - & \textbf{100.0} & 75.0 & -  & \textbf{95.0} & 65.0  \\ \hline

\multirow{4}{*}{18} 
&   8 & 63.4 & 49.9 & 40.8  & 20.0 & 15.0 & 15.0 & 100.0 & 95.0 & 55.0  \\ \cline{2-11}
&  16 & 71.0 & 55.8 & 45.8 & 45.0 & 35.0 & 35.0 & 100.0 & 95.0 & 65.0  \\ \cline{2-11}
&  32 & 85.3 & 67.0 & 53.0 & 60.0 & 55.0 & 40.0  & 100.0 & 95.0 & 65.0   \\ \cline{2-11}
&  64 & - & - & - & -  & - & - & -  & - &  - \\ \hline

\multirow{4}{*}{24} 
&   8 & 67.8 & 52.7 & 39.2 & 20.0  & 20.0 & 15.0 &  100.0 & 90.0 & 50.0  \\ \cline{2-11}
&  16 & 77.4 & 60.7 & 48.7 &  45.0 & 40.0 & 40.0 &  100.0 & 95.0 & 65.0 \\ \cline{2-11}
&  32 & - & 73.7 & 58.3 & -  & 60.0 & 45.0 &  - & 95.0 &  65.0 \\ \cline{2-11}
&  64 & - & - & - & -  & - & - & -  & - &  - \\ \hline

\end{tabular}}
\end{adjustbox}

\end{table}
% -----------------------------------------------------

As discussed in Section~\ref{subsec:model_precision}, the configuration with \(n\!=\!6\) and \(d_{\text{model}}\!=\!64\) under 8-bit quantization achieved optimal RMSE. According to Table~\ref{tab:model_resource}, this configuration barely fit the FPGA, utilizing at least 89.5\% of all types of resources. For 6-bit quantization, the optimal RMSE was attained with the configuration of \(n\!=\!12\) and \(d_{\text{model}}\!=\!64\). Although its test RMSE is 7.33\% higher than our best 8-bit quantized model, its LUTs utilization is 14.2\% lower, while DSPs utilization increases by 5\%. Among all 4-bit quantized models, the two most precise ones were too large to be deployed. We chose the model with the third-best RMSE performance (5.474), which has a 56.13\% higher RMSE than our best 8-bit quantized model. However, its RMSE is only 0.63\% higher than that of the 8-bit quantized model in~\cite{becnel2022tiny}. This model, with the configuration of \(n\!=\!12\) and \(d_{\text{model}}\!=\!32\), consumes only 46.8\% LUTs, 40\% BRAM, and 65\% DSPs. It represents a feasible compromise between resource utilization and precision, making it suitable for resource-constrained applications.

% -----------------------------
\begin{table*}[!htb]
\footnotesize
\centering
\caption{Performance Comparison on Spartan-7 XC7S15 FPGA}
\label{tab:model_hardware}
\setlength{\tabcolsep}{1.2mm}{
\begin{tabular}{|c|c|c|c|c|c|c|c|c|}
\hline
\multirow{2}{*}{\begin{tabular}[c]{@{}c@{}}Configs.\\ ($n$, $d_\text{model}$, $b$)\end{tabular}} & \multirow{2}{*}{RMSE} & \multirow{2}{*}{\begin{tabular}[c]{@{}c@{}}Clock\\ Frequency(MHz)\end{tabular}} & \multirow{2}{*}{\begin{tabular}[c]{@{}c@{}}Clock\\ Cycles\end{tabular}} & \multirow{2}{*}{Time(ms)$^\dagger$} & \multicolumn{3}{c|}{Power(mW)$^\dagger$}  & \multirow{2}{*}{Energy(mJ)} \\ \cline{6-8}
&  &  &   & & \multicolumn{1}{c|}{Static} & Dynamic & Total & \\ \hline

(6, 64, 8)  & 3.506  & 100  & 282974 & 2.82 & 31 & 44 & 75 & 0.212 \\ \hline
(12, 64, 6) & 3.763  & 125  & 575696 & 4.61 & 31 & 48 & 79 & 0.364 \\ \hline
(12, 32, 4) & 5.474  & 125  & 166394 & 1.33 & 31 & 32 & 63 & 0.084 \\ \hline
\multicolumn{9}{l}{\small $\dagger$  The numerical estimates obtained from GHDL and Vivado exhibit 2\% variance in time,} \\
\multicolumn{9}{l}{\small  \quad  and 5\% variance in power when compared to the actual hardware measurements.} 
\end{tabular}}
\vspace{-15pt}
\end{table*}

% -----------------------------

\subsection{Timing Analysis}
\label{subsec:timing}

Focusing on the three candidate configurations identified in Section~\ref{subsec:resource_util}, we analyzed the timing performance of these FPGA-deployable models. As detailed in the third column of Table \ref{tab:model_hardware}, the clock frequency of the 8-bit quantized model was limited to 100 MHz. In contrast, models with lower bitwidths operate at frequencies up to 25\% higher, supporting the expectation that fewer bits expedite computation due to simplified logic. However, decreasing the bitwidth from 6-bit to 4-bit did not increase the frequency, likely due to reduced DSPs engagement and subsequent increases in logic delay.

The fourth column of Table \ref{tab:model_hardware} presents the number of clock cycles required per inference, which is influenced by model configuration (\(n\), \(d_{\text{model}}\)). Comparing Rows 2 and 3, doubling \(n\) resulted in 2.03$\times$ more clock cycles per inference. Similarly, comparing Rows 3 and 4, halving \(d_{\text{model}}\) led to 3.46$\times$ fewer clock cycles per inference. The fifth column of Table \ref{tab:model_hardware} outlines the model inference time, which depends on the clock frequency and the number of clock cycles. Notably, our 4-bit quantized model demonstrated the shortest inference time, while the 6-bit quantized model exhibited the longest.

% 282974,0000000 8-bit => 282974*(1/100/1000000)=0.00282974*1000 = 2.82
% 575696,0000000 6-bit => 575696*(1/125/1000000)=0.004605568*1000 = 4.61
% 166394,0000000 4-bit => 166394*(1/125/1000000)=0.001331152*1000 =1.33

% ---------------------------------------------------------------------------
\subsection{Power and Energy Consumption}

The power estimates from Vivado are also summarized in Table \ref{tab:model_hardware}. Interestingly, the 6-bit quantized model exhibited slightly higher power consumption than the 8-bit quantized model, likely due to its increased clock frequency. However, reducing the bitwidth of computation is generally expected to conserve power, a theory validated by the 4-bit quantized model, which demonstrated the lowest total power consumption at 63 mW among the three configurations.

Energy consumption, as shown in the last column of Table \ref{tab:model_hardware}, is calculated based on power usage and inference time. The 4-bit quantized model was the most energy-efficient, owing to its shorter inference time and lower power consumption. In contrast, the 6-bit quantized model underperformed the others, incurring the highest costs in both power and time. Notably, despite the 4-bit quantized model exhibiting a 56.13\% higher RMSE than the 8-bit model, it is 2.12 times faster and 2.52$\times$ more energy-efficient. This efficiency highlights the potential advantages of deploying the 4-bit quantized model on smaller FPGAs, where resource constraints are more pronounced. Conversely, our 8-bit quantized model remains a suitable choice when precision is paramount.

%%%%%%%%%%%%%%%%%%%%%%%%%%%%%%%%%%%%%%%%%%%%%%%%%%%%%%%%%%%%%%%%%%%%%%%%%%%%%%%%
\section{Related Work}
\label{sec:related_work}

Research on quantizing Transformer-based architectures has been extensive, particularly in the domains of NLP~\cite{kim2021bert} and CV~\cite{li2023vit}. These models, known for their substantial computational demands, are primarily deployed on cloud-based servers and edge servers equipped with GPUs. However, relevant studies in TS, especially time-series forecasting, remain underexplored. Becnel et al.~\cite{becnel2022tiny} implemented an 8-bit quantized Transformer on a low-power microcontroller unit (MCU) for time-series forecasting, achieving a power consumption of 23 mW but with an inference time of 176 ms. In contrast, our 4-bit quantized Transformer accelerator on an embedded FPGA achieves up to 132.33$\times$ faster inference with only a 0.63\% increase in test loss. Despite a 2.74$\times$ increase in power consumption, it consumes 48.19$\times$ less energy, indicating that integrating an FPGA-based accelerator with the MCU could be beneficial.

Previous works~\cite{yamini2023hardware, sobakinskikhoptimizing} have explored implementing Transformer models on FPGAs for time-series applications, but their target platforms are server-grade or edge-grade FPGAs, which are unsuitable for embedding into IoT devices. In contrast, we chose the Xilinx Spartan-7 FPGA as our target platform. Although resource-constrained, it offers a balanced solution in terms of speed and energy efficiency for deploying Transformer models.

%%%%%%%%%%%%%%%%%%%%%%%%%%%%%%%%%%%%%%%%%%%%%%%%%%%%%%%%%%%%%%%%%%%%%%%%%%%%%%%%
\section{Conclusion and Future Work}
\label{sec:conclusion_future_work}

In this study, we implemented an FPGA-friendly Transformer model using software-hardware co-design to balance model precision, resource utilization, timing, power and energy. We adopted QAT in the PyTorch framework to ensure model precision and conducted low-bit integer-only inference simulations prior to accelerator generation. Furthermore, our approach utilized VHDL templates and automatic generation scripts to facilitate the seamless translation of trained quantized models into FPGA-ready accelerators. Through hardware validation, we confirmed the effectiveness of our approach.

In future work, we plan to extend our approach to mixed-precision quantization. Additionally, we aim to optimize the hardware implementation further to improve energy efficiency, thereby advancing the sustainability of Transformer models for time-series forecasting in AIoT systems.

\vspace{-5pt}
% ---------------------------------------------------------------------------
\bibliographystyle{IEEEtran}
\bibliography{reference}
\end{document}